\documentclass[runningheads,a4paper]{llncs}

\usepackage{amssymb}
\usepackage{subfig}
\usepackage{color}
\usepackage{graphicx}
\usepackage[hyphens]{url}
\usepackage[hidelinks]{hyperref}
\hypersetup{breaklinks=true}
\usepackage{url}
\urldef{\mailsa}\path|{noura.al-moubayed, toby.breckon,| 
\urldef{\mailsb}\path|peter.matthews, stephen.mcgough }@durham.ac.uk|

\begin{document}

\mainmatter  

\title{SMS Spam Filtering using Probabilistic Topic Modelling and Stacked Denoising Autoencoder}

\titlerunning{SMS Spam Filtering using Probabilistic Topic Modelling and Deep Learning}

%
%
\author{Noura Al Moubayed
\and Toby Breckon \and Peter Matthews \and A. Stephen McGough}
\authorrunning{Al Moubayed, Breckon, Matthews, and McGough }
%
\institute{School of Engineering and Computing Sciences,\\
Durham University, DH1 3LE Durham, UK\\
\mailsa\\
\mailsb\\
\url{}}

\toctitle{Lecture Notes in Computer Science}
\tocauthor{Authors' Instructions}
\maketitle

\begin{abstract}
In This paper we present a novel approach to spam filtering and demonstrate its applicability with respect to SMS messages. Our approach requires minimum features engineering and a small set of labelled data samples. Features are extracted using topic modelling based on latent Dirichlet allocation, and then a comprehensive data model is created using a Stacked Denoising Autoencoder (SDA). Topic modelling summarises the data providing ease of use and high interpretability by visualising the topics using word clouds. Given that the SMS messages can be regarded as either spam (unwanted) or ham (wanted), the SDA is able to model the messages and accurately discriminate between the two classes without the need for a pre-labelled training set. The results are compared against the state-of-the-art spam detection algorithms with our proposed approach achieving over 97\% accuracy which compares favourably to the best reported algorithms presented in the literature.

\end{abstract}


\section{Introduction}
Short Messaging Service (SMS) applications are the most widely used applications on smart phones \cite{smith2015smartphone}  where 97\% of surveyed users in the report used SMS at least once during the survey. People worldwide were expected to send 8.3 trillion text messages on 2013 alone \cite{PortioReport}. The large volume of SMS traffic is opening up an opportunity for spammers to move from email to SMS spamming \cite{GSMA11}.

Prior research has shown that the most effective approach for spam filtering is to perform the threat analysis on the message content level\cite{SMSSurvey}. The SMS problem is in principle very similar to email spam filtering \cite{review3,TiagoEmail}. However, SMS differs mainly due to the nature of SMS messaging itself: 1) SMS is capped at 160 characters. 2) Users normally write an idiosyncratic language subset with abbreviations, bad spelling, SMS slang, and internet acronyms. Despite this most filters use standard feature extraction methods such as direct N-gram character-based and word-based  tokenisation \cite{gomez2006content}. Supervised and unsupervised machine learning techniques are commonly trained using a collection of labelled messages of spam and non-spam (usually referred to as ham) \cite{SMSSurvey}. The trained model is then used to predict labels of previously unseen messages. 

In this work we use a recently developed text mining method, that of probabilistic topic modelling  \cite{steyvers2007latent}, to extract the hidden topics that are statistically related to SMS. Topic modelling has the advantage of handling seamlessly and robustly any text size \cite{steyvers2007latent}. The topics generated per SMS are then used by an unsupervised deep learning approach, stacked denoising auto-encoders (SDA)  \cite{vincent2010stacked}, to build a data model. A novel onset detection approach based on the built SDA model is then used to increase separation between ham and spam and finally a Fisher's linear discriminate analysis (FDA)\cite{scholkopft1999fisher} is used to classify data into spam and ham. The results achieved using this approach are comparable with the best reported in the literature.


\section{SMS Spam Filtering}

The first step in a machine learning based SMS spam filter is feature extraction/engineering. The classifier must effectively utilise these features for discrimination of spam and ham. This is by no means a unique problem for spam filtering, however, the limited available text per SMS makes the feature space sparse. This means that the samples, from the input space, are fewer and further apart, thus significantly reducing the data that the classifier has to work with \cite{SMSSurvey}. Hidalgo et al \cite{gomez2006content} suggested the use of different features including: normalised words, character bi- and tri-grams and word bi-grams. A novel approach based on Stylometry, i.e. the statistical analysis of linguistic style, was presented in \cite{sohn2009contribution}, with the goal of identifying spam message from the style by which those messages were written. In their review of email spam filtering, \cite{review3} reported that the bag of words was the most common feature used in the literature. However, they argue that the greatest disadvantage of this approach was that the features are fixed and can not be updated as the data changes and the nature of spam threat changes. The extracted features tend to be high dimensional requiring some sort of feature selection, or dimensionality reduction techniques \cite{SMSSurvey,sohn2009contribution,gomez2006content}.

After the features are extracted and selected, the machine learning method can be trained to classify the available data into spam and ham. Early work suggested the use of both supervised machine learning methods, e.g. SVM \cite{xiang2004filtering}, and unsupervised methods, e.g. k-NN \cite{healy2004assessment}. Hidalgo et al \cite{gomez2006content} evaluated a number of spam filtering methods and concluded that SVMs are the most suitable classification approaches. As the number of spam samples in any dataset is much smaller than that of ham samples, any classifier must take this into consideration otherwise there is a serious risk of over-fitting the model to one class (usually ham). To address this issue a Bayesian approach to a Naive Bayes based classifier was used \cite{jie2010bayesian}. This approach penalises false positives more ensuring balanced performance for ham and spam and higher spam precision.

\section{Methods}
\label{sec:methods}
The most commonly used methods for SMS feature extraction suffer from three main disadvantages: 1) the number of resulting features are usually high requiring the use of a feature selection method  2) the features can be very sparse due to the limited size of SMS 3) the selected features are normally hard-coded in the system and hence are very hard to adapt to emerging spam patterns. To address these issues we have opted to use probabilistic topic modelling \cite{steyvers2007latent}, a text mining technique that models latent patterns in the messages, that models latent patterns in the text. This approach automatically identifies topics within a set of messages and assigns each message to a set of topics. The approach only requires the maximum number of topics to be set. The messages are distributed among a small number of topics minimising the effect of sparsity. The most importantly topic modelling can work adaptively. Topic modelling also requires only basic pre-processing steps: tokenisation and stop words removal.

Due to the limited availability of labelled training data, unsupervised learning is the most realistic approach for real-life applications. Here we use an unsupervised deep neural network: stacked denoising autoencoders \cite{vincent2010stacked} (SDA). SDAs are usually pre-trained using an unsupervised approach and then a supervised method is used for fine-tuning. In our approach we only utilise the pre-trained stage with the reconstruction error of a data sample given the model used as a surrogate measure of how well the sample is represented by the model and hence is exploited to identify outliers (e.g. spam).

\subsection{Probabilistic Topic Modeling}

Topic modelling \cite{steyvers2007latent} is a text mining tool that can identify latent text patterns in a documents contents, handling large volumes of corpuses regardless of the size of the individual documents. It describes, in statistical terms, how words in documents are generated based on a pre-defined number of topics using a statistical sampling technique. A commonly used topic modelling method is Latent Dirichlet Allocation (LDA) \cite{blei2003latent}. In LDA the documents are represented by a pre-defined number of topics where each topic is a hidden variable characterised by a nominal distribution over a fixed dictionary. LDA  represents each document as a mixture of different topics with prior assumptions about their distribution. A topic may occur in different documents with a different probability and  a word may occur in several topics with a different probabilities. A complete description of LDA can be found in \cite{blei2003latent}. Let $V$ be a vocabulary consisting of a set of words, $T$ is a set of $k$ topics and $n$ documents of arbitrary length. For every topic $z$ a distribution $\varphi_z$ on $V$ is sampled from a known probability distribution (Dirichlet function \cite{johnson2002continuous}). Gibbs sampling is normally used for inference in LDA. LDA estimates the distribution $p(z|w)$ for $z \in T^P$ , $w \in V^P$ where $P$ denotes the set of word positions in the documents. 





\subsection{Stacked Denoising Autoencoder}

The main advantage of the unsupervised deep learning is the utilisation of the previously considered useless masses of unlabelled data that are easy to obtain in order to achieve better understanding of emerging patterns in the data. Unsupervised deep learning is capable of extracting high level feature representations of complex structured data outperforming approaches based on handcrafted features \cite{bengio2012unsupervised}.   

An autoencoder ($AE$) consists of a visible input layer, and a hidden layer. During learning the AE goes through two phases: 1) construct phase which maps the input data into the hidden layer 2) reconstruct phase which maps back the hidden layer's data into the input layer. The model converges when the reconstruction error between input and output is minimum. $AE$ normally use tied (constrained) weights for regularisation  \cite{bengio2012unsupervised}. This constrains the parameter search space and reduces the number of parameters to learn: $W$, also known as the weight matrix. The constructed representation of the input $x$, can be defined as $y=S(W x +a)$ and the reconstructed representation of $y$ can be defined as  $z =S(W^\prime y +b)$,
where $W^\prime$ is the transpose of $W$, and $S(\bullet)$ is a sigmoid function ($S(x)= \frac{1}{1+e^{-x}}$). The reconstruction error is measured using squared error:$
 L(x,z) = \parallel x-z\parallel^2$. The model is then optimised to find the $W$ that minimises $L$. 

To avoid over-fitting, i.e. learning the identify function, and reduce information redundancy in the input features we use a Denoising Autoencoder (DA) \cite{vincent2010stacked}. DA  is a stochastic version of the AE that corrupts the input data by adding noise, allowing for more variance in the input space and hence better generalisation of the model. In this paper we adopt the Masking Noise corruption forcing a fraction of the input layer units (chosen randomly) to have a weight of $0$.

\begin{figure}
\centering
\includegraphics[width=10cm, height=7cm]{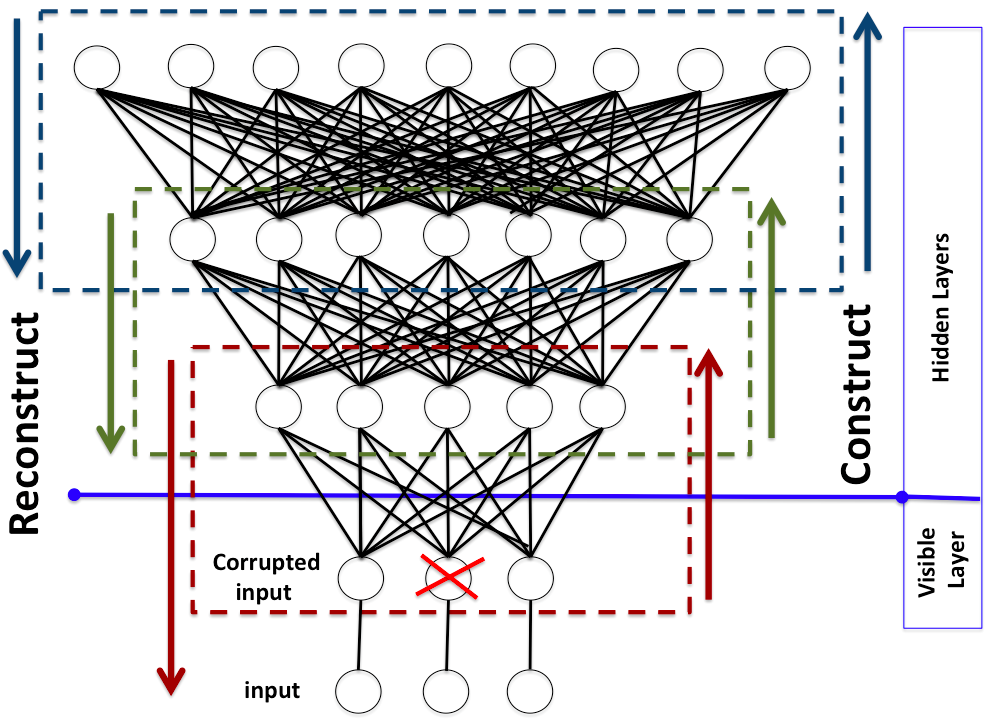}
\caption{A sample SDA model architecture. The crossed node in the input layer represents data corruption.}
\label{fig:sda}
\end{figure}

Stacked Denoising Autoencoder (SDA) is the deep version of a single DA, where the output of one DA is the input to the following one. The network is then trained layer by layer. Fig. \ref{fig:sda}  illustrates the SDA architecture. The arrows indicate the direction of information flow. During construction the data flows from the input layer up in the hierarchy to the top layer. For reconstruction the data flows back from the top through the hidden layers down to the input layer where the reconstructed data is compared with the input data and the overall reconstruction error (RE) is calculated. 

\subsection{Outlier Detection}

Reconstruction error is a measure of how well SDA models the presented sample at the input layer. A high RE suggests poor modelling of the input sample while a small RE is an indication of accurate representation of the input. RE among layers is only used during unsupervised pre-training to optimise the model parameters. In this work we utilise overall RE in a novel way as a measure for detecting outliers (i.e. spam). As the majority of available data is ham SDA will model them more accurately than spam. In other words, spam will have higher RE than ham making it easier to discriminate the two sets (Fig \ref{fig:rec} B) using simple linear classifiers like FDA \cite{scholkopft1999fisher}.

\section{Experiments and Results}

\label{sec:results}

The SMS spam data was collected and first presented in \cite{almeida2011contributions}. The data contains 5574 messages: 747(13.40\%) labelled as spam and 4827 (86.60\%) labelled as ham.

\begin {table}[h]
\caption {Classification Results} \label{grandResults} 
\begin{center}
 \begin{tabular}{||c| c| c| c| c||} 
 \hline
 Classifier & SC\% & BH\% & Acc\% & MCC\% \\ [0.5ex] 
 \hline\hline
  TM+SDA & 85.59 & 0.62  & 97.51 & 0.899\\ 
 \hline
 Logistic Reg. + tok2 & 95.48 & 2.09 & 97.59 & 0.899\\ 
 \hline
 SVM + tok1 & 83.10 & 0.18 & 97.64 & 0.893 \\
 \hline
 Boosted NB + tok2 & 84.48 & 0.53 & 97.50 & 0.887 \\
 \hline
 SMO + tok2 & 82.91 & 0.29 & 97.50 & 0.887 \\
 \hline
 Boosted C4.5 + tok2 & 81.53 & 0.62 & 97.05 & 0.865 \\ 
 \hline
 MDL + tok1 & 75.44 & 0.35 & 96.26 & 0.826\\  
 \hline
PART + tok2 & 78.00 & 1.45 & 95.87 & 0.810 \\  
 \hline
Random Forest + tok2 & 65.23 & 0.12 & 95.36 & 0.782\\ 
 \hline
C4.5 + tok2 & 75.25 & 2.08 & 95.00 & 0.770 \\  
 \hline
Bern NB + tok1 & 54.03 & 0.00 & 94.00 & 0.711\\  
 \hline
MN TF NB + tok1 & 52.06 & 0.00 & 93.74 & 0.697 \\  
 \hline
MN Bool NB + tok1 & 51.87 & 0.00 & 93.72  & 0.695\\  
 \hline
1NN + tok2 & 43.81 & 0.00 & 92.70  & 0.636\\  
 \hline
Basic NB + tok1 & 48.53 & 1.42 & 92.05  &0.600\\  
 \hline
 Gauss NB + tok1 & 47.54 & 1.39 & 91.95 & 0.594 \\  
 \hline
 1Flex NB + tok1 & 47.35 & 2.77 & 90.72  &0.536\\  
 \hline
 Boolean NB + tok1 & 98.04 & 26.01 & 77.13 & 0.507\\  
 \hline
 3NN + tok2 & 23.77 & 0.00 & 90.10 & 0.462\\  
 \hline
 EM + tok2 & 17.09 & 4.18 & 85.54 &0.185 \\  
 \hline
 TR & 0.00 & 0.00 & 86.95 & - \\  
 \hline 
\end{tabular}
\end{center}
\end{table}

First the text content of the messages is tokenised, and stop words are removed. No stemming is applied to the data as this may affect the interpretability of the topic modelling results. The pre-processed text is then used to build a dictionary and bag of words which are passed to LDA to generate the topic model. Ham contains a wide range of topics that are irrelevant to the discrimination between spam and ham. Hence, only data labelled as spam was employed in building the topic model. A maximum of 60 topics were used. This was the optimal value identified after varying the maximum number of topics between 10 and 100. After the model was built all the messages (ham and spam) were passed to the model producing a 60-feature vector per message, where a feature $i$ is the probability of that message $j$ contains topic $i$.

SDA uses an input layer of 60 units with two hidden layers of 100, and 150 units respectively. All units use sigmoid activation functions with the learning rate is set to 0.1 and corruption rate of 30\%. The learning algorithm runs for 100 epochs. The learnt model is then used to calculate RE for each message, followed by FDA classification. To properly evaluate the performance of the methods a 10-fold cross validation approach was used. For each fold the training data was used to build a topic model and generate the feature vectors for training and testing data. SDA is built using the training features and REs are used to train an FDA which was then tested on RE of the testing set. This process is repeated 10 times and the average accuracies are reported.

\begin{figure}[h]
\centering
\includegraphics[width=12cm]{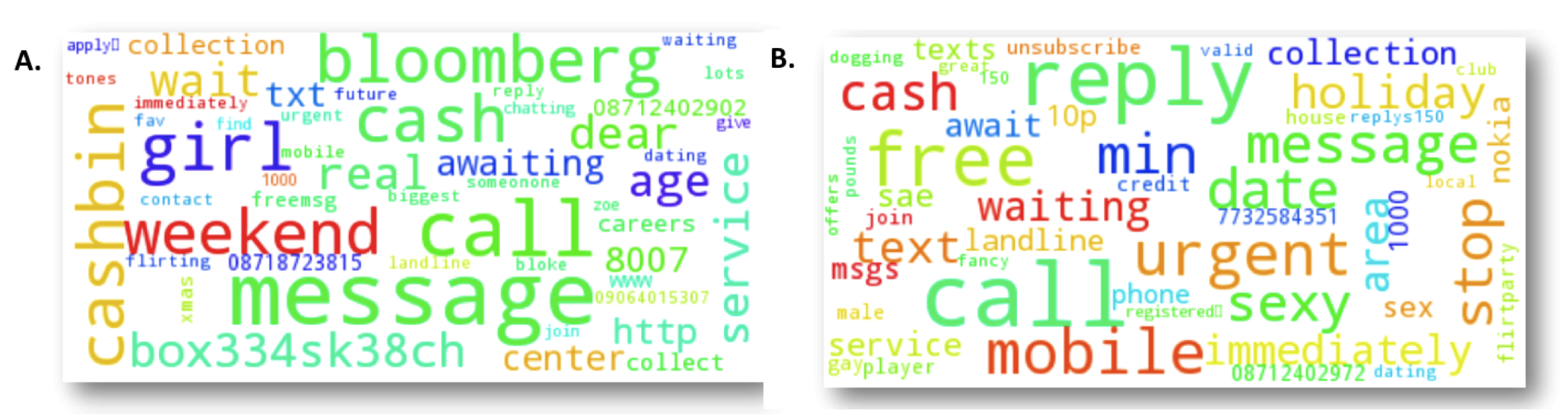}
\caption{The result of the application of LDA topic modelling on spam data (A. is topic 2 and B. is topic number 59). The size of the word is proportional to the probability of that word belonging to the topic. }
\label{fig:topics}
\end{figure}

One of the major advantages of topic modelling is the ability to visualise the topics and interpret their meaning using a word cloud presentation. Figure \ref{fig:topics} demonstrates the word cloud of two distinct topics generated by the same topic model. It is clear that some words are joint between the two topics but with different probabilities. 

\begin{figure}[h]
\centering
\includegraphics[width=12cm, height=4cm]{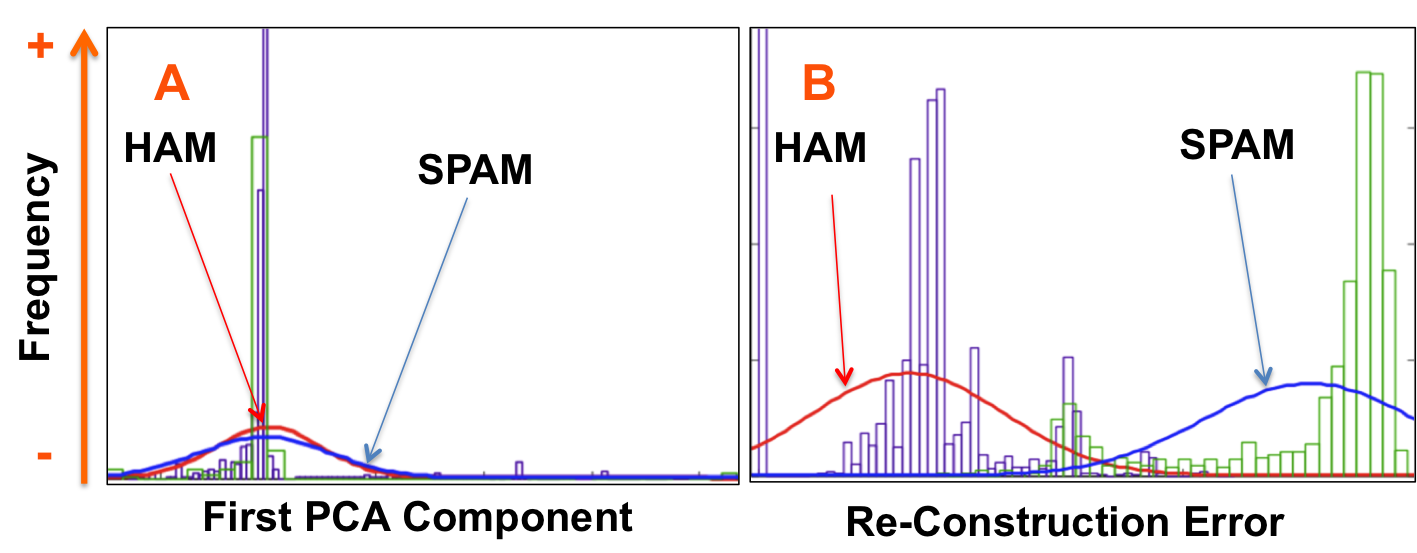}
\caption{A. Distribution of first PCA component of both ham and spam data. B. Distribution of reconstruction errors for ham and spam. }
\label{fig:rec}
\end{figure}

Figure \ref{fig:rec} plots the histogram and fitted Gaussian probability density function for ham and spam. The figure clearly shows a high separability between the two classes using SDA, while a principal component analysis (PCA) approach fails. It shows the ability of SDA to build a model for ham data resulting in small REs, while it does not fit the spam data as well resulting in higher REs.

Our cross-validated approach results in F-score = 90.13 $\pm$ 3.4 (mean $\pm$ standard deviation), Precision= 95.47 $\pm$ 1.9, and Recall = 85.58 $\pm$ 6.0. However to keep with the evaluation metrics reported in the literature \cite{almeida2011contributions} we also report the overall cross validated classification accuracy (Acc\%), the Spam Caught accuracy (SC \%), Blocked Ham accuracy (BH\%), and Mathews Correlation Coefficient (MCC\%). Table \ref{grandResults} presents our results as TM+SDA along with the commonly used methods in the literature \cite{almeida2011contributions} ordered by MCC\%. 

\subsection{Conclusions}

This paper presents a novel approach for SMS spam filtering using recent advances in text mining and unsupervised outlier detection based on deep learning. Topic modelling is proposed as the feature extraction method which tackles several disadvantages of the state-of-the-art methods. By modelling the abstract topics responsible of generating the text within a given message, a limited number of features can be used eliminating the need for feature selection. The model also reduces the sparsity in the input space making it easier for the classifier to decode the data. The model itself is adaptive so it can cope with newly emerging data samples without the need for a  major redesign of the system. This, along with the ease of use and interpretability the topic model approach offers, allows us to argue that this approach has significant advantage in many application areas.

SDA was presented as an unsupervised technique to model the extracted topic modelling features. SDA is demonstrated here to successfully separate between ham and spam using the structure in the data alone  without the need for any labelling. The novelty of our approach is to use reconstruction errors produced by SDA to increase separability between ham and spam. FDA classifier trained on RE is then very effective in classifying the two classes. The accuracy achieved by the proposed system is comparable to the best results reported in the literature (using logistic regression (LR)). Although LR scores higher than ours on spam caught, it scores worse on ham blocked. 

As SDA is completely unsupervised, the approach is scalable to large unlabelled data sets and requires only a small subset to be labelled for FDA training. 



\Urlmuskip=0mu plus 1mu\relax
\bibliographystyle{splncs03}
\bibliography{Noura_ICANN16}

\end{document}